%% file: 00_main.tex
\documentclass[10pt,twocolumn,letterpaper]{article}

\usepackage[pagenumbers]{cvpr} 

\usepackage{graphicx}
\usepackage{amsmath}
\usepackage{amssymb}
\usepackage{booktabs}
\usepackage{times}
\usepackage{epsfig}
\usepackage{nccmath}
\usepackage{pifont}
\usepackage{multirow}
\usepackage{array}
\usepackage{enumerate}
\usepackage{enumitem}
\usepackage{subcaption}
\usepackage{caption}
\usepackage{rotating}
\usepackage[misc]{ifsym}
\usepackage{url}
\usepackage[title]{appendix}
\newcolumntype{L}[1]{>{\raggedright\let\newline\\\arraybackslash\hspace{0pt}}m{#1}}
\newcolumntype{C}[1]{>{\centering\let\newline\\\arraybackslash\hspace{0pt}}m{#1}}
\newcolumntype{R}[1]{>{\raggedleft\let\newline\\\arraybackslash\hspace{0pt}}m{#1}}

\usepackage{xcolor}
\definecolor{cgreen}{RGB}{34, 139, 34}
\newcommand{\cmark}{\color{cgreen}\ding{51}}%
\newcommand{\xmark}{\color{red}\ding{55}}%
\AtBeginDocument{\setlength\abovedisplayskip{3pt}}
\AtBeginDocument{\setlength\belowdisplayskip{2pt}}
\usepackage[pagebackref,breaklinks,colorlinks]{hyperref}
\newcommand*\samethanks[1][\value{footnote}]{\footnotemark[#1]}

\usepackage[capitalize]{cleveref}
\crefname{section}{Sec.}{Secs.}
\Crefname{section}{Section}{Sections}
\Crefname{table}{Table}{Tables}
\crefname{table}{Tab.}{Tabs.}

\def\cvprPaperID{5943} 
\def\confName{CVPR}
\def\confYear{2022}

\begin{document}

\title{Learning Part Segmentation through Unsupervised Domain Adaptation \\from Synthetic Vehicles}

\author{Qing Liu\textsuperscript{1}\thanks{Corresponding author: qingliu.research@gmail.com}, Adam Kortylewski\textsuperscript{1}, Zhishuai Zhang\textsuperscript{1}, Zizhang Li\textsuperscript{2}\thanks{This work is done during internship at Johns Hopkins University.}\\
Mengqi Guo\textsuperscript{3}\samethanks[2], Qihao Liu\textsuperscript{1}, Xiaoding Yuan\textsuperscript{4}\samethanks[2], Jiteng Mu\textsuperscript{5}\samethanks[2], Weichao Qiu\textsuperscript{1}, Alan Yuille\textsuperscript{1}\\
\textsuperscript{1}Johns Hopkins University\quad\textsuperscript{2}Zhejiang University\\
\textsuperscript{3}Beihang University\quad\textsuperscript{4}Tongji University\quad\textsuperscript{5}University of California, San Diego
}

\maketitle

\begin{abstract}
\vspace{-0.3cm}
Part segmentations provide a rich and detailed part-level description of objects. However, their annotation requires an enormous amount of work, which makes it difficult to apply standard deep learning methods. In this paper, we propose the idea of learning part segmentation through unsupervised domain adaptation (UDA) from synthetic data. We first introduce UDA-Part, a comprehensive part segmentation dataset for vehicles that can serve as an adequate benchmark for UDA\footnote{\url{https://qliu24.github.io/udapart/}}. In UDA-Part, we label parts on 3D CAD models which enables us to generate a large set of annotated synthetic images. We also annotate parts on a number of real images to provide a real test set. Secondly, to advance the adaptation of part models trained from the synthetic data to the real images, we introduce a new UDA algorithm that leverages the object's spatial structure to guide the adaptation process. Our experimental results on two real test datasets confirm the superiority of our approach over existing works, and demonstrate the promise of learning part segmentation for general objects from synthetic data. We believe our dataset provides a rich testbed to study UDA for part segmentation and will help to significantly push forward research in this area.
\end{abstract}

\input{01_introduction}
\input{02_related_work}
\input{03_method}
\input{04_experiments}
\input{05_conclusion}

{\small
\bibliographystyle{ieee_fullname}
\bibliography{egbib}
}

\clearpage
\newpage
\onecolumn
\begin{appendices}
\input{06_supplementary}
\end{appendices}
\twocolumn

\end{document}

%% file: 01_introduction.tex
\vspace{-0.4cm}
\section{Introduction}
\vspace{-0.2cm}
\begin{figure}[t]
\centering
\includegraphics[width=0.85\linewidth]{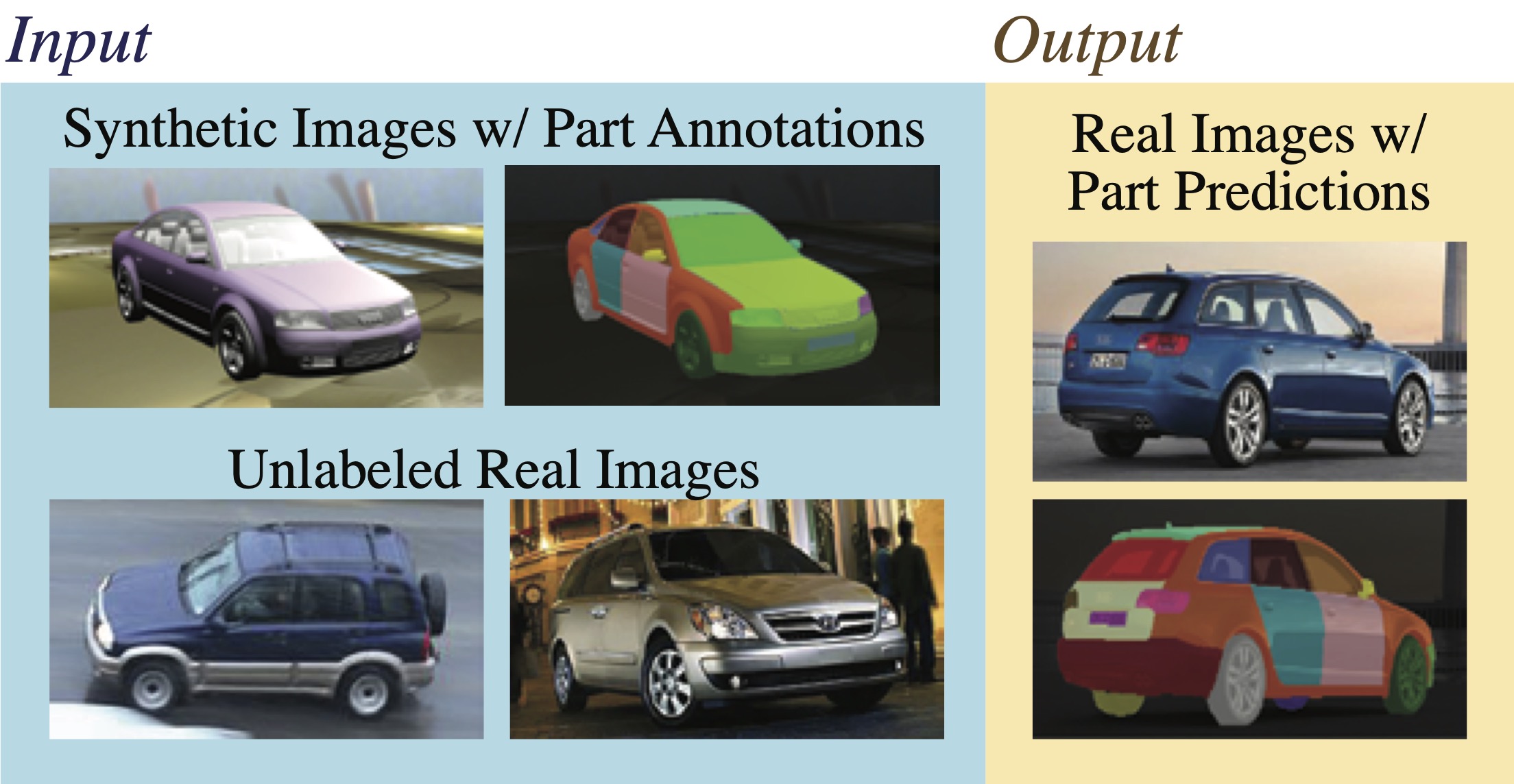}
\vspace{-0.3cm}
\caption{An illustration of learning part segmentation through unsupervised domain adaptation (UDA) from synthetic vehicles. Based on part annotation on 3D CAD models, we propose to use UDA to learn from large-scale labeled synthetic samples and unlabeled real images, and the goal is to make accurate part segmentation predictions on real test images.}
\label{fig:uda}
\vspace{-0.4cm}
\end{figure}

Part-based object representations are of key importance for many computer vision tasks such as object recognition~\cite{zhang2014part,chen2014detect,sun2013learning,azizpour2012object}, pose estimation~\cite{khan2017head,xiang2014beyond,dong2014towards,yang2011articulated}, action detection~\cite{wang2012discriminative}, and scene understanding~\cite{shi2016data,topfer2013efficient,stark2011fine}. Currently, part-based approaches often represent objects as a set of sparse keypoints, because these are easy to annotate in large-scale datasets for training deep neural networks.
By contrast, part segmentations provide a richer and more detailed part-level object description. Instead of recognizing specific parts sparsely on the object (e.g., keypoint or part detection), part segmentation gives a complete description of an object by assigning every pixel belonging to the object one and only one part label. This is a lot more challenging task and requires a much greater annotation effort.

Given their recent success, deep learning methods have dominated the studies of computer vision, including object segmentation~\cite{long2015fully,chen2017deeplab,cheng2020panoptic}. However, these deep models usually require a large amount of annotated training data to achieve satisfying performance. Existing part segmentation datasets mostly contain only a small number of images~\cite{chen2014detect,thomas2008using}, or define only a small number of parts per object category~\cite{yi2016scalable,thomas2008using}, or focus on a single object category, such as humans~\cite{gong2018instance,gong2017look,liang2015deep,yamaguchi2012parsing} and faces~\cite{CelebAMask-HQ,khan2017head,le2012interactive}. These limitations inhibit effective training of standard deep segmentation networks and have largely impeded the development of computer vision models that leverage part information. By contrast, 3D CAD models are available for many different objects, and, once annotated, can be used to generate large-scale part segmentation datasets automatically.

In this work, we propose to solve part segmentation on general objects by learning from synthetic data (Figure~\ref{fig:uda}) and address the problem in two steps. In the first step, we introduce UDA-Part, a comprehensive part segmentation dataset that can serve as an adequate benchmark for UDA. UDA-Part is composed of $21$ 3D CAD models from $5$ vehicle categories. For each category, we define a fine-grained set of parts which are consistently annotated across all CAD models of the corresponding category. Based on these CAD models and their part annotations, we are able to render a large-scale synthetic image dataset with automatically generated part segmentation ground-truth. These synthetic data are sufficient to train deep neural networks and may also be used for model evaluation or diagnosis. To evaluate how the models trained from synthetic data perform on the real images, we also label parts on $200$ real images collected from PASCAL3D+~\cite{xiang2014beyond} and include them as the target test set in UDA-Part.

Secondly, we introduce a new unsupervised domain adaptation (UDA) algorithm for part segmentation. UDA has been explored for image classification~\cite{kang2019contrastive,pei2018multi}, keypoint detection~\cite{mu2020learning,zhou2018unsupervised}, and semantic segmentation~\cite{kang2020pixel,zou2019confidence}, where it achieves satisfactory results on real images with little annotation cost. To further advance UDA performance on part segmentation, we introduce Geometric-Matching Guided domain adaptation (GMG). GMG conducts cross-domain geometric matching based on a global transformation function between real and synthetic images. The function puts a smoothness constraint on the matching and thus adaptively preserves the spatial relations between the parts. Once an optimal match is found, GMG transfers the synthetic labels to the real images and retains the high-confidence results as pseudo-labels for a joint training process. In short, GMG makes part-relation-aware adaptation by explicitly using the object structure depicted in the synthetic samples. In our experiments, GMG outperforms other UDA baselines for part segmentation on both the UDA-Part real test images and the PascalPart~\cite{chen2014detect} test set.

In summary, our main contributions are:
\vspace{-0.2cm}
\begin{enumerate}[wide,itemsep=0mm]
    \item We propose to learn part segmentation for general objects through unsupervised domain adaptation (UDA) from synthetic data.
    \item We introduce a new part segmentation dataset for vehicles called UDA-Part which can serve as a comprehensive benchmark for part segmentation through UDA.
    \item We introduce a new UDA algorithm for part segmentation called Geometric-Matching Guided domain adaptation (GMG), which leverages the object's spatial structure to guide the adaptation and achieves superior results.
\end{enumerate}
\vspace{-0.4cm}

%% file: 02_related_work.tex
\section{Related Work}
\vspace{-0.2cm}
\begin{figure*}[t]
\centering
\includegraphics[width=0.9\linewidth]{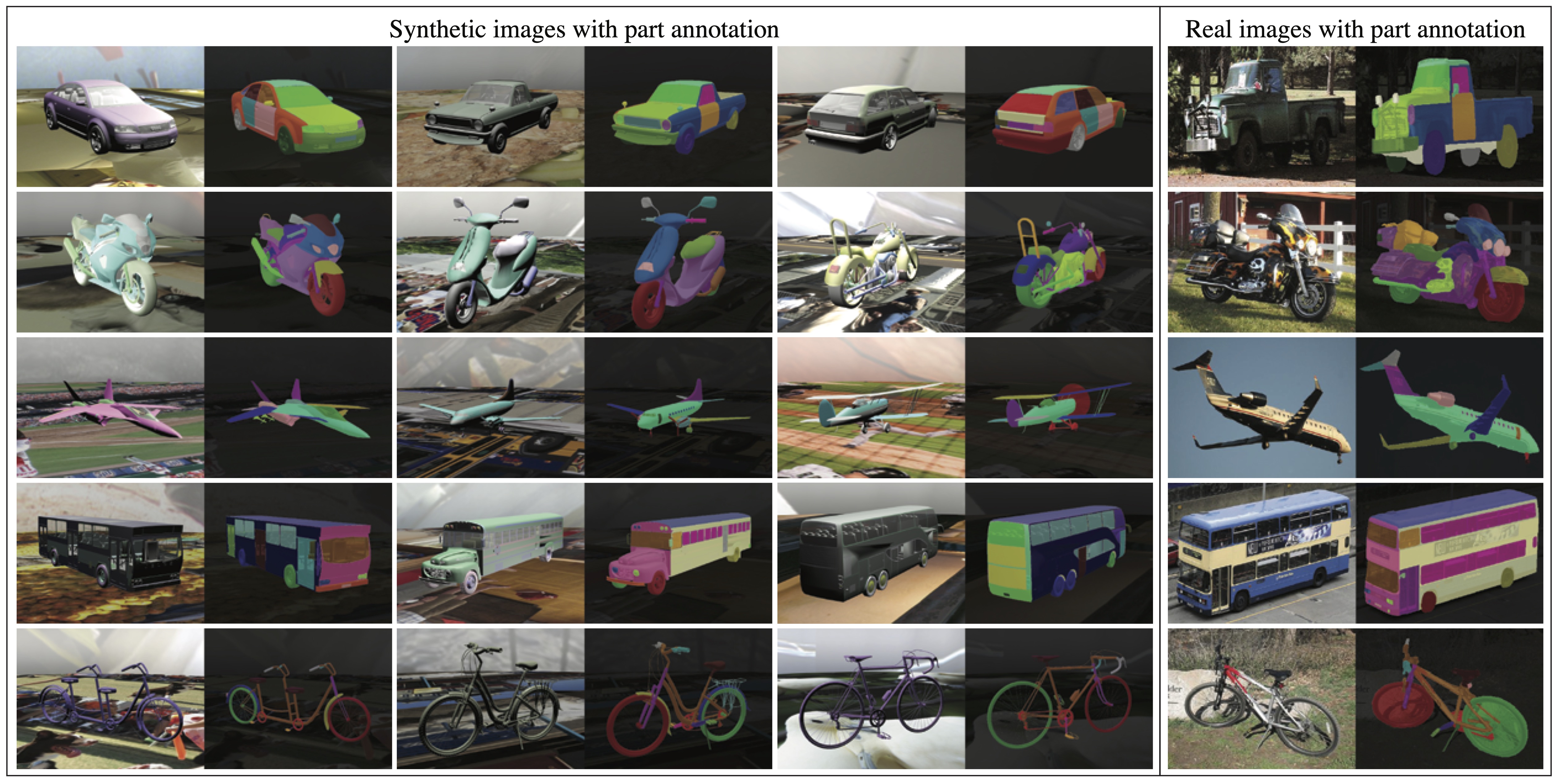}
\vspace{-0.3cm}
\caption{An overview of our UDA-Part dataset. UDA-Part is a comprehensive part segmentation dataset for five vehicle categories. Each category is composed of several 3D CAD models/prototypes. By labeling parts on these 3D CAD models, we render a large-scale synthetic image dataset with automatically generated part segmentation ground-truth (left panel). To benchmark unsupervised domain adaptation methods, we also label parts on a moderate number of real images for testing (right panel). }
\label{fig:overview}
\vspace{-0.4cm}
\end{figure*}

\textbf{Part segmentation.}
Both rigid~\cite{song2017embedding,lu2014parsing} and non-rigid objects~\cite{gong2018instance,liang2015deep,wang2015semantic,yamaguchi2012parsing} have been studied in part segmentation, where structure-based methods, e.g., compositional models, are widely used~\cite{xia2017joint,xia2016pose,wang2015joint,wang2015semantic,lu2014parsing}. Architecture-wise, both fully convolutional network (FCN)~\cite{xia2016zoom,wang2015joint} and long short-term memory (LSTM)~\cite{liang2017interpretable,liang2016semantic} have been studied. Many works explore the use of auxiliary tasks (e.g., pose estimation) to boost part segmentation and get promising results~\cite{fang2018weakly,nie2018mutual,xia2017joint,gong2017look}. Moreover, 3D information such as 3D geometric features~\cite{song2017embedding} and depth~\cite{shotton2011real} could also be embedded into the models to improve the performance. Beyond single-object parsing, multi-object parsing is recently proposed and studied ~\cite{michieli2020gmnet,zhao2019multi}. While all these methods require pixel-level annotations on real images, some works start to use synthetic data to advance human parsing~\cite{kanazawa2018end,varol2017learning}, but they are limited to human object. In this work, we aim to solve part segmentation on general objects by learning from synthetic vehicles.

\textbf{Datasets with part annotations for rigid objects.}
ImageNetPart~\cite{wang2015unsupervised} provides bounding box annotation for parts on $6$ vehicle categories. PASCAL3D+~\cite{xiang2014beyond} includes keypoint annotation on $12$ rigid object categories. CarFusion~\cite{reddy2018carfusion} and ApolloCar3D~\cite{song2019apollocar3d} contain keypoint annotation for cars in street scenes.
For 3D object part recognition, PartNet~\cite{mo2019partnet} provides hierarchical part annotations on 3D models covering $24$ object categories, most of which are indoor furniture and none is a vehicle. Yi et al.~\cite{yi2016scalable} label parts on 3D models selected from $16$ categories in ShapeNetCore~\cite{shapenet2015}, while their definition of part is coarse and the average number of parts per category is less than $4$.
In the context of image part segmentation, PascalPart~\cite{chen2014detect} has been widely studied. It includes $20$ categories but provides limited number of samples and parts. MVP~\cite{liu2020beyond} is recently introduced to provide detailed part segmentation labels for cars in the wild. An earlier dataset ETHZ~\cite{thomas2008using} provides annotations of $5$ parts on $141$ wheelchair images and $6$ parts on $139$ car images, which is not adequate for deep network training. 
PartImageNet \cite{he2021partimagenet} is a very recent large scale dataset consisting of $158$ classes with up to $5$ parts annotated per class, requiring several hundred hours of annotation. 
Our aim is to build a large-scale part segmentation dataset for generic objects efficiently using 3D computer graphics models.

\textbf{Learning from synthetic data.}
Synthetic data generated by computer graphics techniques are effective for model diagnosis~\cite{johnson2017clevr,zhang2018unrealstereo} and have boosted performance in many real-world application domains~\cite{mu2020learning,kortylewski2019analyzing,tremblay2018training,varol2017learning,handa2016understanding,dosovitskiy2015flownet}. However, the domain shift between synthetic data and real-world data limits the improvement. To overcome this, domain adaptation is proposed~\cite{patel2015visual}. Maximum Mean Discrepancy (MMD) and its kernel variants~\cite{sun2016deep,tzeng2015simultaneous,long2015learning,tzeng2014deep} have been studied to reduce the difference between source and target domain distributions.
In the context of unsupervised domain adaptation (UDA) for semantic segmentation, self-training is one of the mainstream research directions~\cite{zou2018unsupervised,zou2019confidence,zhang2019category,li2020content}, while there has also been an increasing interest in using style transfer~\cite{murez2018image,sankaranarayanan2018learning,hoffman2018cycada,wu2018dcan,li2019bidirectional,chang2019all} or feature alignment~\cite{hoffman2016fcns,tsai2018learning} to encourage domain-wise marginal distribution matching. Other methods adopt category-aware feature alignment or local contextual feature similarity~\cite{luo2019taking,zhang2019category,dong2020cscl,huang2020contextual}.
We promote the idea of using UDA to solve part segmentation and leverage object structure to guide the knowledge transfer from synthetic to real.

\textbf{Geometric matching.}
Geometric matching aims at finding spatial correspondences among images belonging to the same category at a fine-grained level. Both hand-engineered descriptors~\cite{kim2013deformable,liu2010sift,berg2005shape} and pre-trained convolutional neural network (CNN) features~\cite{ham2017proposal,ufer2017deep,yang2017object,taniai2016joint} are explored in early works. Recent progress is made in trainable image descriptors~\cite{novotny2017anchornet,kim2017dctm,kim2017fcss} and trainable geometric models~\cite{chen2019arbicon,rocco2018end,rocco2017convolutional,han2017scnet}. However, they have only explored geometric matching for images in the same domain. Bai et al.~\cite{bai2019semantic} demonstrate that pre-trained CNN features can be effectively used to find sparse spatial correspondence between synthetic images and real images. Zhou et al.~\cite{zhou2016learning} use 3D cycle consistency to learn dense correspondence between real-to-real and real-to-synthetic pairs. However, their method does not consider the global structure nor constrain it with a geometric transformation function, thus the result is less competitive.
In this work, we explore cross-domain geometric matching without supervision and integrate it into our part segmentation framework.
\vspace{-0.2cm}


%% file: 03_method.tex
\section{The UDA-Part Dataset}
\vspace{-0.2cm}

In this section, we introduce UDA-Part, a part segmentation dataset for vehicles that provides detailed annotations on both synthetic images for training and real images for testing (Figure~\ref{fig:overview}). UDA-Part can serve as a comprehensive benchmark for part segmentation through UDA.

\vspace{-0.15cm}
\subsection{Data Generation}
\vspace{-0.2cm}
We build UDA-Part in five steps. Examples of the annotations on the synthetic images and the real test images are shown in Figure~\ref{fig:overview}. More details (e.g., CAD models, part list, etc.) could be found in the supplementary material.

\textbf{(1) Select 3D CAD models.} UDA-Part is composed of $21$ 3D CAD models from $5$ vehicle categories: \emph{car}, \emph{motorbike}, \emph{aeroplane}, \emph{bus}, and \emph{bicycle}. Each CAD model represents a common prototype (i.e., subtype) of the category it belongs to. For example, for the category \emph{bicycle}, the CAD models are different bicycle subtypes such as \emph{utility}, \emph{sports}, \emph{road}, and \emph{tandem}. These can effectively represent the structural variability of bicycle objects. In total, we select $21$ CAD models to be included and annotated in UDA-Part.

\textbf{(2) Define list of parts to be studied.} We take references from existing vehicle part datasets~\cite{chen2014detect,thomas2008using} and Wikipedia~\cite{enwiki:1010693301,enwiki:998640799,enwiki:995736929} to determine the part list for each category. Comparing with the part list in PascalPart ~\cite{chen2014detect}, we make more fine-grained level definitions. For example, in PascalPart, the category \emph{car} has a part ``back side''. We instead distinguish ``back windshield'', ``tail light'', ``back bumper'', and ``trunk''. The fine-grained definition used here makes it possible to merge the part list and map to other coarse lists defined in existing datasets.

\textbf{(3) Annotate parts on 3D CAD models.} We adopt the Blender~\cite{blender} plugin built by Kim et al.~\cite{kim2022learn} to perform per-mesh part labeling on the 3D CAD models. The plugin allows the user to assign a label to a group of selected meshes and save the results to a JSON file. We include a quality control step to ensure each surface mesh is assigned one and only one part label. Annotating all CAD models in UDA-Part takes roughly $300$ working hours.

\textbf{(4) Render synthetic images with part annotations.} We use Blender~\cite{blender} as our renderer to generate the synthetic images with part segmentation ground-truth. Following previous work~\cite{prakash2019structured,tremblay2018training}, we randomize the render parameters (e.g., viewpoint, lighting, object texture, etc.) to enable nuisance factor control and facilitate domain generalization. We generate $8000$ synthetic images with resolution $2048\times 1024$ for each 3D CAD model and split the training and test set with a ratio of $3:1$, resulting in a total of $126000$ images for training and $42000$ for testing.

\textbf{(5) Annotate parts on real test images.} We manually label part segmentations on $200$ real vehicle images ($40$ images per category) for testing. The images are selected from PASCAL3D+ dataset~\cite{xiang2014beyond} to contain objects with different subtypes and evenly distributed viewpoints. We use the VGG Image Annotator (VIA)~\cite{dutta2019vgg} to label the parts on the images, which takes about $150$ working hours.

\vspace{-0.15cm}
\subsection{Dataset Comparison}
\vspace{-0.2cm}
\begin{figure}[t]
\centering
\setlength{\tabcolsep}{1mm}
\includegraphics[width=0.86\linewidth]{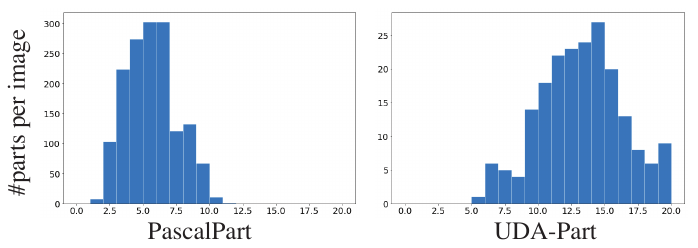}
\vspace{-0.4cm}
\caption{Histogram of number of parts per image in PascalPart~\cite{chen2014detect} test split and UDA-Part real test images. The x-axis represents count of images.}
\label{fig:compare_cnt}
\vspace{-0.3cm}
\end{figure}

In Table~\ref{tab:dataset}, we compare UDA-Part with existing datasets that provide vehicle part segmentation labels. UDA-Part contains the largest number of parts per category, and its annotation on both synthetic images and real images makes it adequate for UDA studies. In Figure~\ref{fig:compare_cnt}, we compare the number of parts per image labeled on PascalPart and UDA-Part real test images. Generally, UDA-Part contains $2$ to $3$ times more annotations per image, making it a more challenging part segmentation benchmark. More comprehensive comparisons between UDA-Part and PascalPart can be found in the supplementary material.

\vspace{-0.15cm}
\section{Geometric-Matching Guided Adaptation}
\vspace{-0.2cm}
\begin{figure*}[t]
\centering
\includegraphics[width=0.9\linewidth]{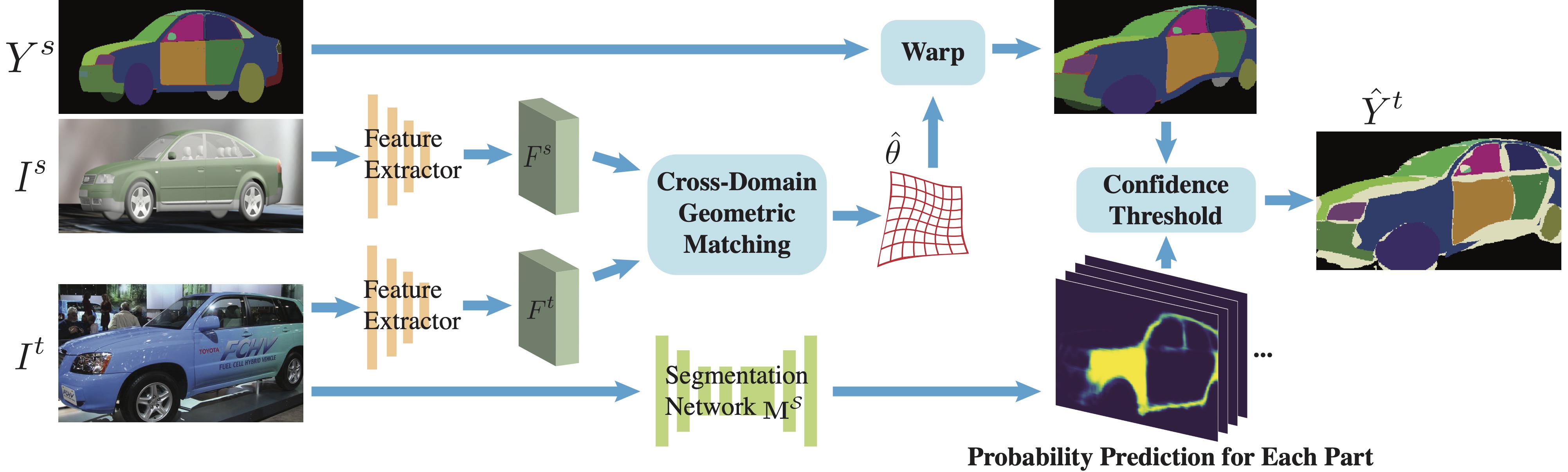}
\vspace{-0.3cm}
\caption{An illustration of Geometric-Matching Guided (GMG) domain adaptation. A pair of synthetic and real images are passed through a feature extractor to get their feature maps, and cross-domain geometric matching is used to estimate a 2-D transformation based on the feature similarities. The transformation is then applied to the segmentation ground-truth of the synthetic image to make it match the parts in the real image. Finally, a confidence threshold is used to filter out unreliable warping results and make high-quality pseudo-labels for the joint training step in unsupervised domain adaptation.}
\label{fig:gmg}
\vspace{-0.4cm}
\end{figure*}

To facilitate studies targeting real-world data applications, we explore unsupervised domain adaptation (UDA) algorithms which enable the models trained on the synthetic data of UDA-Part to perform well on the real test images. In this section, we introduce our proposed Geometric-Matching Guided domain adaptation (GMG) approach. The key steps of GMG are illustrated in Figure~\ref{fig:gmg} and will be discussed in detail in the following.

\vspace{-0.15cm}
\subsection{Preliminaries for Part Segmentation and UDA}\label{sec:prelim}
\vspace{-0.2cm}
We start with the preliminaries for part segmentation and unsupervised domain adaptation (UDA). In our work and all baseline experiments, we assume the category label is known, such that the part segmentation models are trained and tested for each object category separately. This assumption is reasonable since many off-the-shelf classifiers are freely available for the objects included in UDA-Part.

\begin{table}[t]
\centering
\scalebox{0.9}{
\small
\setlength{\tabcolsep}{1mm}
\begin{tabular}{|@{\hspace{2\tabcolsep}}L{16mm}|C{14mm}|C{14mm}|C{12mm}|C{12mm}|}
\hline
                          & \begin{tabular}[c]{@{}c@{}} UDA-Part \\ (Ours) \end{tabular} & \begin{tabular}[c]{@{}c@{}} PascalPart \\ \cite{chen2014detect} \end{tabular} & \begin{tabular}[c]{@{}c@{}} ETHZ \\ \cite{thomas2008using} \end{tabular} & \begin{tabular}[c]{@{}c@{}}SNPart \\ \cite{yi2016scalable} \end{tabular} \\ \hline
3D Models             & \cmark            & \xmark         & \xmark       & \cmark       \\ \hline
Syn. Img.          & \cmark            & \xmark         & \xmark        & \xmark       \\ \hline
Real Img.               & \cmark            & \cmark        & \cmark       & \xmark      \\ \hline
Avg. \#Parts per Cat. &  $24$             & $9$          & $6$        & $4$       \\ \hline
\end{tabular}}
\vspace{-0.35cm}
\caption{Comparison of UDA-Part with existing datasets that have part segmentation labels on vehicles.}
\label{tab:dataset}
\vspace{-0.65cm}
\end{table}

Similar to semantic segmentation, a part segmentation model $\mathbf{M}$ can be formulated as a mapping function from the image domain to the output label domain: $\mathbf{M}:I\rightarrow Y$, which predicts a pixel-wise category label $Y\in \{1,\dots,C\}^{H\times W}$, where $H$ and $W$ denote the image size, $C$ is the total number of part categories. 

In UDA, the data are usually collected from two domains: source domain $\mathcal{S}$ (i.e., synthetic data) and target domain $\mathcal{T}$ (i.e., real data). During training, we have access to the labeled training samples $(I^s, Y^s)$ from $\mathcal{S}_{\text{train}}$ and unlabeled training samples $I^t$ from $\mathcal{T}_{\text{train}}$. The goal is to train a model $\mathbf{M}$ that can predict accurately on the test samples in $\mathcal{T}_{\text{test}}$. 
Most existing approaches start with training a \emph{Source-Only} model $\mathbf{M}^\mathcal{S}$ on the labeled source data. For segmentation models with softmax output, the cross-entropy loss is widely used during the optimization process:
\begin{align*}
    \mathcal{L}_{ce}(I^s, Y^s)=-\sum_{i=1}^H\sum_{j=1}^W\sum_{c=1}^{C} y^s_{(i,j),c}\log p_{(i,j)}(c|I^s;\mathrm{w}),
\end{align*}
where $(i,j)$ are the pixel coordinates in $I^s$, $c$ is the category index, $y^s_{(i,j),c}\in\{0,1\}$ is entry in the one-hot vector of the ground-truth label, i.e., $\forall (i,j)$, $\sum_c y^s_{(i,j),c}=1$, and $p_{(i,j)}(c|I^s;\mathrm{w})$ is the predicted category probability based on the model parameters $\mathrm{w}$.

Generally, $\mathbf{M}^\mathcal{S}$ has limited generalization capability and does not perform well on target samples due to domain discrepancy between the real and synthetic data. One common approach in UDA is using $\mathbf{M}^\mathcal{S}$ to generate pseudo-labels $\hat{Y}^t$ on $I^t$, which enables a joint training on $\mathcal{S}_{\text{train}}$ and $\mathcal{T}_{\text{train}}$ based on the following loss function:
\begin{align*}
    \mathcal{L}({ \mathcal{S}_{\text{train}}},{ \mathcal{T}_{\text{train}}})=\sum_{ \mathcal{S}_{\text{train}}}\mathcal{L}_{ce}(I^s, Y^s)+\lambda\sum_{ \mathcal{T}_{\text{train}}}\mathcal{L}_{ce}(I^t, \hat{Y}^t),
\end{align*}
where $\lambda$ balances the loss between source and target domain. The joint training process encourages the learning of domain-invariant features and shared decision boundaries.

Different UDA methods have been proposed to find reliable pseudo-labels $\hat{Y}^t$~\cite{mu2020learning,zhang2019category,zou2018unsupervised}, select a relevant subset of $\mathcal{S}_{\text{train}}$~\cite{li2020content}, or add regularization terms and adversarial losses~\cite{li2019bidirectional,zou2019confidence} for the joint training process. However, since most of these methods are designed for semantic segmentation, none of them takes advantage of the object's spatial structure to predict parts. On the contrary, our proposed Geometric-Matching Guided domain adaptation (GMG) explicitly uses the structural relation between parts to generate pseudo-labels for joint training.

\vspace{-0.15cm}
\subsection{Cross-Domain Geometric Matching}
\vspace{-0.2cm}
In GMG, we use cross-domain geometric matching to find an optimal global transformation that can be used to transfer segmentation labels from synthetic to real. Cross-domain geometric matching has been previously explored for few-shot learning and correspondence learning~\cite{bai2019semantic, zhou2016learning}, but either with sparse alignment or strong supervision. Here we use it to facilitate the domain adaptation for part segmentation at dense pixel-level and require no labels on the real images in the whole training process.

Specifically, cross-domain geometric matching aims to find spatial correspondence between a pair of synthetic and real images $(I^s,I^t)$. To achieve this, we optimize a global transformation function $\mathbf{W}_\theta$ that matches the two images $I^s$ and $I^t$ based on their feature similarities. $\mathbf{W}_\theta$ puts smoothness constraint on the matching, so it is able to preserve the spatial relations between parts. After the matching, the transformation is applied to transfer the synthetic label $Y^s$ to the real image $I^t$ as a pseudo-label $\hat{Y}_t$. In the following, we first assume that the input image pair is given and the objects in both images belong to the same prototype and have a similar viewpoint. We then discuss how to search such input pairs in an unsupervised manner. 
Note that cross-domain geometric matching is only used to generate pseudo-labels for the joint-training. At test time, neither a paired input nor a geometric matching is necessary in our framework.

Given $(I^s,I^t)$ with similar appearances, we first use CNN convolutional layers to extract their feature maps. The output $F^s$ and $F^t$ are $h\times w\times d$ tensors, which can be interpreted as $h\times w$ grids of $d$-dimensional local features $f_{(i,j)}\in \mathbb{R}^d$. The similarity between two feature vectors in $F^s$ and $F^t$ can be measured by cosine similarity:
\begin{align*}
    \phi(f^s_{(i,j)},f^t_{(k,l)}) = \frac{f^s_{(i,j)} \cdot f^t_{(k,l)}}{||f^s_{(i,j)}||_2 ~ ||f^t_{(k,l)}||_2}.
\end{align*}
To emphasize the similarities are measured for features sampled from different domains, we use $(i,j)$ to denote spatial coordinates in $I^s$, and $(k,l)$ for spatial coordinates in $I^t$.

We then define a 2D geometric transformation function $\mathbf{W}_{\theta}: \mathbb{R}^2 \rightarrow \mathbb{R}^2$ so the spatial correspondence between $I^t$ and $I^s$ could be found by $(k',l') = \mathbf{W}_{\theta}(k, l)$, where $\theta$ denotes the transformation parameters, $(k',l')$ are the corresponding coordinates of $(k,l)$ in $I^s$.

The quality of a geometric transformation can be measured by the sum of feature similarities at corresponding coordinates:

\begin{align*}
\Phi_\theta(F^s, F^t) &= \sum_{(k,l)}\phi(f^s_{\mathbf{W}_{\theta}(k,l)},f^t_{(k,l)}),
\end{align*}
and our goal is to find the best parameters $\hat{\theta}$ such that:
\begin{align*}
\hat{\theta} & =\arg\max_{\theta}\Phi_\theta(F^s, F^t).
\end{align*}

In practice, we follow Rocco et al.~\cite{rocco2018end} and use a spatial transformer layer~\cite{jaderberg2015spatial} to implement the warping, which makes $\Phi$ differentiable w.r.t $\theta$. 
Note that we just optimize the transformation parameters based on the feature similarities. The CNN backbone is fixed in this step.

\textbf{Unsupervised selection of input pairs.} For each real training image $I^t\in\mathcal{T}_{\text{train}}$, we perform a grid search over the viewpoint and prototype in the synthetic images and select the best one based on $\Phi_{\hat{\theta}}$. More specifically, we first build a pool of prototypical synthetic images by selecting samples from each prototype with $24$ diverse viewpoints (i.e., azimuth angles sampled from $\{0,30,60,\dots,330\}$ and elevation angles sampled from $\{5,20\}$). Then, we perform geometric matching for $I^t$ and each $I^s$ in this pool. The synthetic image that achieves the highest $\Phi_{\hat{\theta}}$ is selected, and its label is warped to infer the pseudo-label for $I^t$.

\vspace{-0.15cm}
\subsection{Confidence Threshold of Pseudo-Labels}
\vspace{-0.2cm}
Given $\hat{\theta}$, we can warp $Y^s$ to $I^t$ and use it as pseudo-supervision for joint training. However, the warping results can contain errors due to the variability of the object shapes and 3D poses, or due to a sub-optimal estimation of the transformation parameters. To correct such errors, in GMG, we use the confidence of the prediction provided by the \emph{Source-Only} model $\mathbf{M}^\mathcal{S}$. Specifically, for spatial coordinates $(k,l)$ in $I^t$, if the corresponding coordinates $\mathbf{W}_{\hat{\theta}}(k,l)$ in $I^s$ have ground-truth label $\bar{c}$, we use the predicted probability of $\bar{c}$ by $\mathbf{M}^\mathcal{S}$ at $(k,l)$ as a confidence score:
\begin{align*}
z_{kl} = p_{(k,l)}(\bar{c}|I^t;\mathrm{w}_\mathcal{S}),
\end{align*}
and threshold $z_{kl}$ with $\gamma$ to obtain the final pseudo-label:
\begin{align*}
    \hat{y}^t_{(k,l)} = \begin{cases}
    y^s_{\mathbf{W}_{\hat{\theta}}(k,l)}, & \text{if } z_{kl}>\gamma \\
    \mathbf{0},              & \text{otherwise},
\end{cases}
\end{align*}
where $\textbf{0}$ denotes vector with $0$ entries everywhere, and is ignored by the cross-entropy loss. Consequently, GMG is able to select high-confidence warping results as pseudo-labels for joint training.

\vspace{-0.3cm}

%% file: 04_experiments.tex
\section{Experiments}
\vspace{-0.1cm}
\begin{table*}[t!]
\centering
\setlength{\tabcolsep}{1mm}
\small
\begin{tabular}{|L{28mm}|C{20mm}|C{20mm}|C{20mm}|C{20mm}|C{20mm}|}
\hline
                 & car ($32$)   & motorbike ($22$) & aeroplane ($23$) & bus ($33$)  & bicycle ($18$)\\ \hline
Source Only  $\mathbf{M}^\mathcal{S}$     & $37.19$ & $15.91$  & $13.71$ &  $19.43$ & $18.01$         \\ \hline
CRST~\cite{zou2019confidence}  &   $39.35$    &    $18.39$    &   $13.38$     &    $21.03$   &    $15.31$     \\
BDL~\cite{li2019bidirectional}  &   $43.60$    &   $17.15$        &   $16.71$  &   $22.95$  &   $17.78$      \\
CAG~\cite{zhang2019category}  &    $48.97$   &     $17.89$      &    $16.06$   &    $25.74$   &    $\textbf{19.50}$  \\
CCSSL~\cite{mu2020learning} & $49.31$ & $21.87$     & $17.50$ &  $24.75$ &    $17.27$     \\ \hline
GMG (Ours)       & $\underline{49.93}$ & $\underline{23.09}$     &  $\underline{17.89}$  &  $\underline{25.78}$ &    $19.07$     \\
GMG (Ours) w/ vp & $\textbf{53.77}$ & $\textbf{23.28}$     & $\textbf{17.98}$  &  $\textbf{26.31}$ &      $\underline{19.09}$   \\ \hline
\end{tabular}
\vspace{-0.3cm}
\caption{Part segmentation results (mIoU) on UDA-Part real test images. The number of parts is denoted in parenthesis next to the category name. ``GMG'' uses unsupervised grid search to find the input pairs, while ``GMG w/ vp'' uses the ground-truth viewpoint of the real training images to reduce matching error. \textbf{Best} and \underline{second best} UDA results are marked accordingly.}
\label{tab:cgpartrst}
\vspace{-0.4cm}
\end{table*}

\begin{table*}[t]
\centering
\setlength{\tabcolsep}{1mm}
\small
\begin{tabular}{|C{10mm}|L{28mm}|C{20mm}|C{20mm}|C{20mm}|C{20mm}|C{20mm}|}
\hline
\multicolumn{2}{|l|}{}     & car ($14$)   & motorbike ($7$) & aeroplane ($7$) & bus ($14$)  & bicycle ($8$)\\ \hline
\multicolumn{2}{|c|}{Fully-Supervised Learning}  & $40.36$ & $38.08$     & $42.47$     & $34.42$ &  $40.57$ \\ \hline
\multirow{7}{*}{UDA}
& Source Only  $\mathbf{M}^\mathcal{S}$    & $14.24$ & $20.48$     & $23.18$     & $16.02$ &  $20.21$ \\ \cline{2-7}
& CRST~\cite{zou2019confidence}  &  $14.44$   &   $24.71$   &  $23.04$      &   $16.56$ &    $22.57$ \\
& BDL~\cite{li2019bidirectional}  &  $19.02$ &   $26.89$    &     $29.08$  &  $17.29$  &   $22.45$  \\
& CAG~\cite{zhang2019category}  &   $18.39$    &    $24.09$       &      $28.60$     &    $17.12$   &     $22.22$    \\
& CCSSL~\cite{mu2020learning}  & $24.23$ & $28.80$     & $32.58$     & $18.59$ &   $22.25$ \\ \cline{2-7}
& GMG (Ours)       & $\underline{25.61}$ & $\underline{29.68}$     & $\underline{33.50}$     & $\underline{19.30}$ &  $\underline{22.91}$ \\
& GMG (Ours) w/ vp & $\textbf{27.59}$ & $\textbf{30.73}$     & $\textbf{33.98}$     & $\textbf{21.20}$ & $\textbf{23.63}$ \\ \hline
\end{tabular}
\vspace{-0.3cm}
\caption{Object part segmentation results (mIoU) on PascalPart test images. The number of parts is denoted in parenthesis next to the category name. ``GMG'' uses unsupervised grid search to find the input pairs, while ``GMG w/ vp'' uses the ground-truth viewpoint of the real training images to reduce matching error. \textbf{Best} and \underline{second best} UDA results are marked accordingly.}
\label{tab:pascalpartrst}
\vspace{-0.4cm}
\end{table*}

In the experiments, different UDA methods are trained using the synthetic data in the source dataset and unlabeled real training images in the target dataset, and are then evaluated on the real test images in the target dataset.

\textbf{Datasets and evaluation metric.}
In all experiments, we use the synthetic samples of the UDA-Part dataset as the source data. UDA-Part provides $30000$ / $24000$ / $24000$ / $24000$ / $24000$ training samples for the object categories car / motorbike / aeroplane / bus / bicycle. 
We evaluate all methods on two target datasets.
In the first set of experiments, \textbf{PASCAL3D+}~\cite{xiang2014beyond} is used as the target dataset, which contains $2763$ / $624$ / $986$ / $548$ / $661$ unlabeled training images for car / motorbike / aeroplane / bus / bicycle. After training, the domain adaptation models are evaluated on $200$ ($40$ per category) real images in the test split, which are selected and annotated as the real test set of UDA-Part.
In the second set of experiments, \textbf{PascalPart}~\cite{chen2014detect} is used as the target dataset. We pre-process PascalPart data by cropping images to contain single object which leads to $538$ / $261$ / $266$ / $221$ / $252$ training images, and $520$ / $255$ / $280$ / $229$ / $263$) test images. Note that the segmentation labels for the PascalPart training images are not used. We train with the dense part labels defined in UDA-Part first and then merge the predictions to PascalPart label space during testing. All models are trained and tested on samples belonging to each vehicle category separately. \textbf{Mean Intersection over Union (mIoU)} is used as the metric for the part segmentation task, where IoU is first computed for each part and then averaged over all parts belonging to the corresponding category.

\textbf{Baseline methods.}
For comparison purposes, we adapt several popular UDA methods from related tasks to object part segmentation. BDL~\cite{li2019bidirectional}, CRST~\cite{zou2019confidence}, and CAG~\cite{zhang2019category} are all methods proposed for semantic segmentation but follow different strategies: BDL uses cycleGAN to reduce pixel-level domain discrepancy and encourage marginal feature alignment; CRST performs self-training with smoothness regularization; while CAG applies adversarial training as initialization and explores category-aware feature alignment during self-training. 
In addition, we test CCSSL~\cite{mu2020learning}, a self-training-based method designed for keypoint detection. CCSSL uses consistency constraints to select reliable pseudo-labels and applies strong data augmentation to improve the model's generalization capability. The code for all baselines was adapted from the original public repositories and is modified to use the same backbone and input size.

\textbf{Implementation details.}
We use DeepLabv3+~\cite{deeplabv3plus2018} as the segmentation network for GMG and all baseline methods. The weights are initialized from an ImageNet~\cite{deng2009imagenet} pre-trained model. We implement our model using Pytorch~\cite{NEURIPS2019_9015} on two TitanX GPUs. Synthetic training images are resized to have a long edge of $800$ pixels while real training images are resized to have a short edge of $224$. For the geometric matching, we use thin plate spline transformation with $25$ anchor points and take the first four convolutional blocks of an ImageNet~\cite{deng2009imagenet} pre-trained VGG16 network~\cite{simonyan2014very} as the feature extractor. The confidence threshold $\gamma$ is set to the $60$th percentile of the scores obtained from all samples in the corresponding category. We implement the pair selection using python with $8$ parallel CPU threads. For a pool of $24$ candidates, it takes $2.1$ seconds per image and roughly $54$ ($195$) minutes for the training set of PascalPart (PASCAL3D+). When the ground-truth viewpoint is given, the matching takes less than $0.4$ second per image, and thus roughly $10$ ($30$) minutes for PascalPart (PASCAL3D+). During joint training, we apply strong augmentations to synthetic images following~\cite{mu2020learning}. The joint training takes $10000$ iterations and the learning rate is fixed at $2.5e-4$. The real loss coefficient $\lambda$ is set to $1.0$ for all ``w/vp'' experiments and $0.1$ for all others. More training details could be found in the supplementary material.

\vspace{-0.15cm}
\subsection{Main Results}
\vspace{-0.2cm}

\begin{figure*}[t]
\centering
\setlength{\tabcolsep}{1mm}
\includegraphics[width=0.9\linewidth]{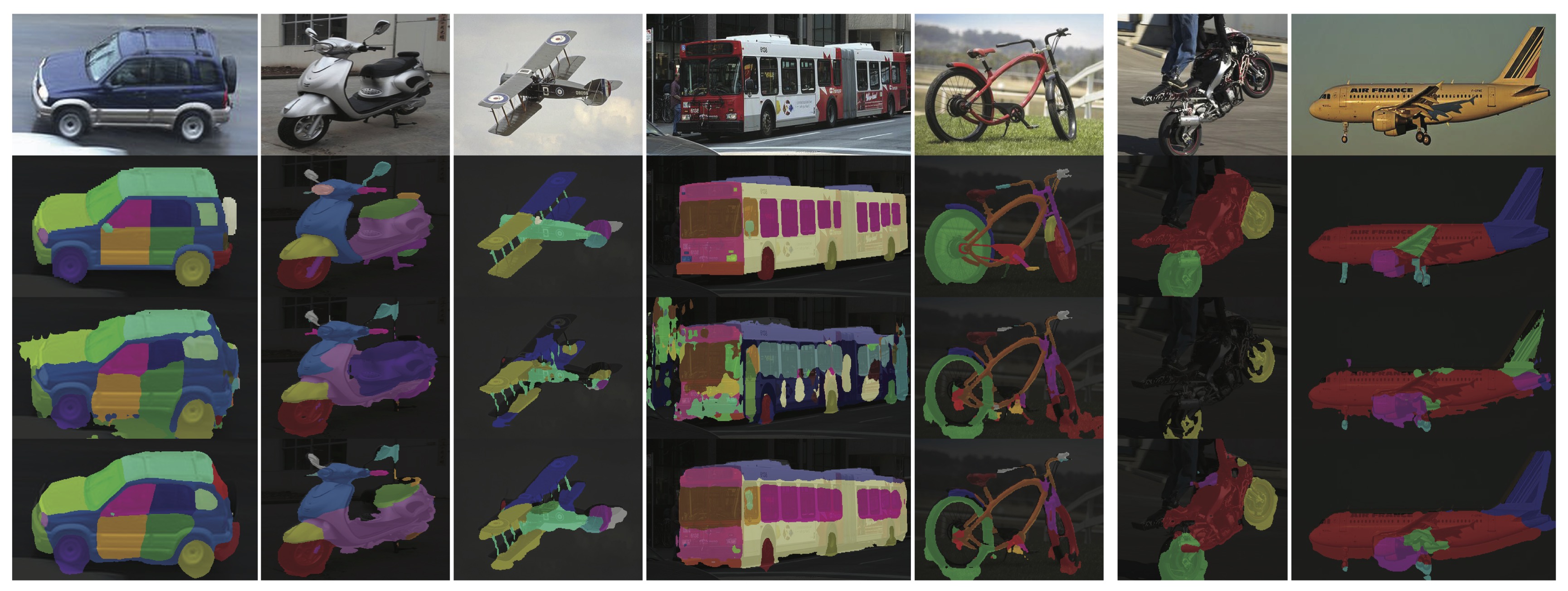}
\vspace{-0.3cm}
\caption{Visualizations of GMG results. We show examples of part segmentation predictions on test images from UDA-Part dataset (column 1-5) and PascalPart dataset~\cite{chen2014detect} (column 6,7). Row 1-4 are real test images, part segmentation ground-truths, \emph{Source-Only} model predictions, and GMG predictions for each case respectively.}
\label{fig:GCG_rst2}
\vspace{-0.5cm}
\end{figure*}

In Table~\ref{tab:cgpartrst}, we report object part segmentation results on the real test images of UDA-Part. In the header row, the number of parts is denoted in parenthesis next to the category name. The \emph{Source-Only} model $\mathbf{M}^\mathcal{S}$ serves as a na\"ive baseline. We observe that the mIoU is generally low, indicating that the domain shift between UDA-Part synthetic images and real-world images indeed disturbs the model performance and domain adaptation is necessary. By looking at the overall performance, segmenting parts on \emph{motorbike}, \emph{aeroplane}, and \emph{bicycle} are more challenging since their parts are often small, with irregular shape and self-occlusion, which can also be observed in Figure~\ref{fig:overview}.

Among the baseline methods, CRST only achieves marginal improvement and may even hurt the performance in some cases, implying that simple smoothness regularization is not working very well on this task. Moreover, since our synthetic images are generated with random texture and background, it is hard to reduce pixel-level domain discrepancy, so the results of BDL are not satisfactory either. CAG achieves more improvement on \emph{car} and \emph{bus}, showing category-aware feature alignment may work well on parts with regular shapes and large areas. CCSSL improves most categories, suggesting part segmentation benefits more from consistency constraints and strong augmentations.

Compared with the baselines, GMG achieves very competitive results, outperforming all other methods in most cases. Furthermore, we introduce a variant of GMG, where pseudo-label errors are reduced by using the ground-truth viewpoints of the real training images during geometric matching (the grid search step only looks for the best prototype). The results from this model are shown in the row ``GMG (Ours) w/ vp'', which show further improvement.

We report object part segmentation results on PascalPart test images in Table~\ref{tab:pascalpartrst}. For this dataset, we include part segmentation results from a DeepLabv3+ network directly trained from PascalPart labels on real images in a fully-supervised manner. These results can be considered as upper-bound for UDA approaches. Compared with UDA-Part, the PascalPart test set seems more manageable for some categories since the parts are more coarse-grained with larger sizes. On the other hand, the images in PascalPart are overall with lower resolution and contain more truncated/occluded objects, making the segmentation task more challenging. Therefore, the result patterns on different categories are not the same as what we observe on UDA-Part test images. Despite the variations in individual test cases, GMG still outperforms all baseline methods on all categories. Similarly, adding ground-truth viewpoints during training consistently improves GMG performance. Comparing with the fully-supervised learning results, we are aware there is still a large room for improvement, indicating a relevant future research direction.

In Figure~\ref{fig:GCG_rst2}, we compare object part segmentation results from GMG and the \emph{Source-Only} model. Generally, GMG is better at recognizing the shape and the boundary of parts. GMG can also eliminate wrong predictions that violate the part relations, implying it gains more knowledge about the object structure. On the other hand, GMG is prone to misclassifying small parts, such as the \emph{mirror} of \emph{car} and the \emph{pedal} of \emph{bicycle}, whose labels are harder to be correctly transferred through geometric matching. We consider it as future work to improve GMG performance on smaller parts.

\vspace{-0.15cm}
\subsection{Ablation and Model Diagnosis}
\vspace{-0.2cm}
\begin{table}[h]
\vspace{-0.2cm}
\centering
\small
\begin{tabular}{|L{15mm}|l|C{10mm}|C{10mm}|C{10mm}|}
\hline
\multicolumn{2}{|l|}{GMG Variants}       & car   & mtbk  & arpl  \\ \hline
\multicolumn{2}{|l|}{Source-Only $\mathbf{M}^\mathcal{S}$}        & $37.19$ & $15.91$ & $13.71$ \\ \hline
\multirow{2}{*}{ \begin{tabular}[c]{@{}l@{}} $+$warped \\ labels \end{tabular}
} & w/o vp & $47.84$ & $20.65$ & $16.25$ \\ \cline{2-5} 
                                & w/ vp  & $53.26$ & $21.94$ & $17.22$ \\ \hline
\multirow{2}{*}{\begin{tabular}[c]{@{}l@{}} $+$conf. \\ thresh. \end{tabular}} & w/o vp & $49.93$ & $23.09$ & $17.89$ \\ \cline{2-5} 
                                & w/ vp  & $53.77$ & $23.28$ & $17.98$ \\ \hline
\end{tabular}
\vspace{-0.3cm}
\caption{Ablation of GMG components. Results on UDA-Part real test images are shown. (mtbk: motorbike; arpl: aeroplane.)}
\label{tab:ablation}
\vspace{-0.3cm}
\end{table}

We quantitatively evaluate the improvement introduced by different components of GMG in Table~\ref{tab:ablation}. Compared with the \emph{Source-Only} model, a joint training based on directly warped synthetic labels leads to a performance gain by leveraging the pseudo-supervision on real samples. However, this method is more sensitive to misalignment caused by wrong geometric matching. Consequently, we observe a larger performance gap between the models that use viewpoint supervision (\emph{w/ vp}) and those without (\emph{w/o vp}). Applying confidence threshold improves the results in all cases, especially when ground-truth viewpoints are not available for real training samples. In summary, we can observe the effectiveness of both components in our model.

In input pair selection, we observe that thresholding $\phi$ can improve viewpoint estimation accuracy, but it leads to reduced number of pseudo-labels for joint training and eventually hurts the segmentation results for categories with less samples, e.g. motorbike. Searching in more azimuth bins indeed improves the viewpoint accuracy but has little positive effect on the final part segmentation mIoU, while searching in more elevation bins makes little difference in both metrics. We also observe GMG is generally not very sensitive to $\gamma$. When we change $\gamma$ from $60$th percentile of the scores to $50$th ($70$th) percentile, the mIoU results for car on PASCAL3D+ changes from $53.77$ to $53.65$ ($53.59$). More details could be found in the supplementary results.

In Figure~\ref{fig:GCG_rst}, we visualize GMG pseudo-label quality and the synthetic samples selected by grid search for the corresponding real images. In the first three examples, the search process successfully finds synthetic images with reasonable prototype and viewpoint. Consequently, geometric matching is performed on image pairs with similar appearances, and generates high-quality pseudo-labels for joint training. In the fourth row, we show a failure case where a wrong (i.e., opposite) viewpoint is selected. Nevertheless, most of the incorrect labels in this case are filtered out by confidence threshold and will not disturb the joint training in this case.

\begin{figure}[h]
\centering
\setlength{\tabcolsep}{1mm}
\vspace{-0.4cm}
\includegraphics[width=0.9\linewidth]{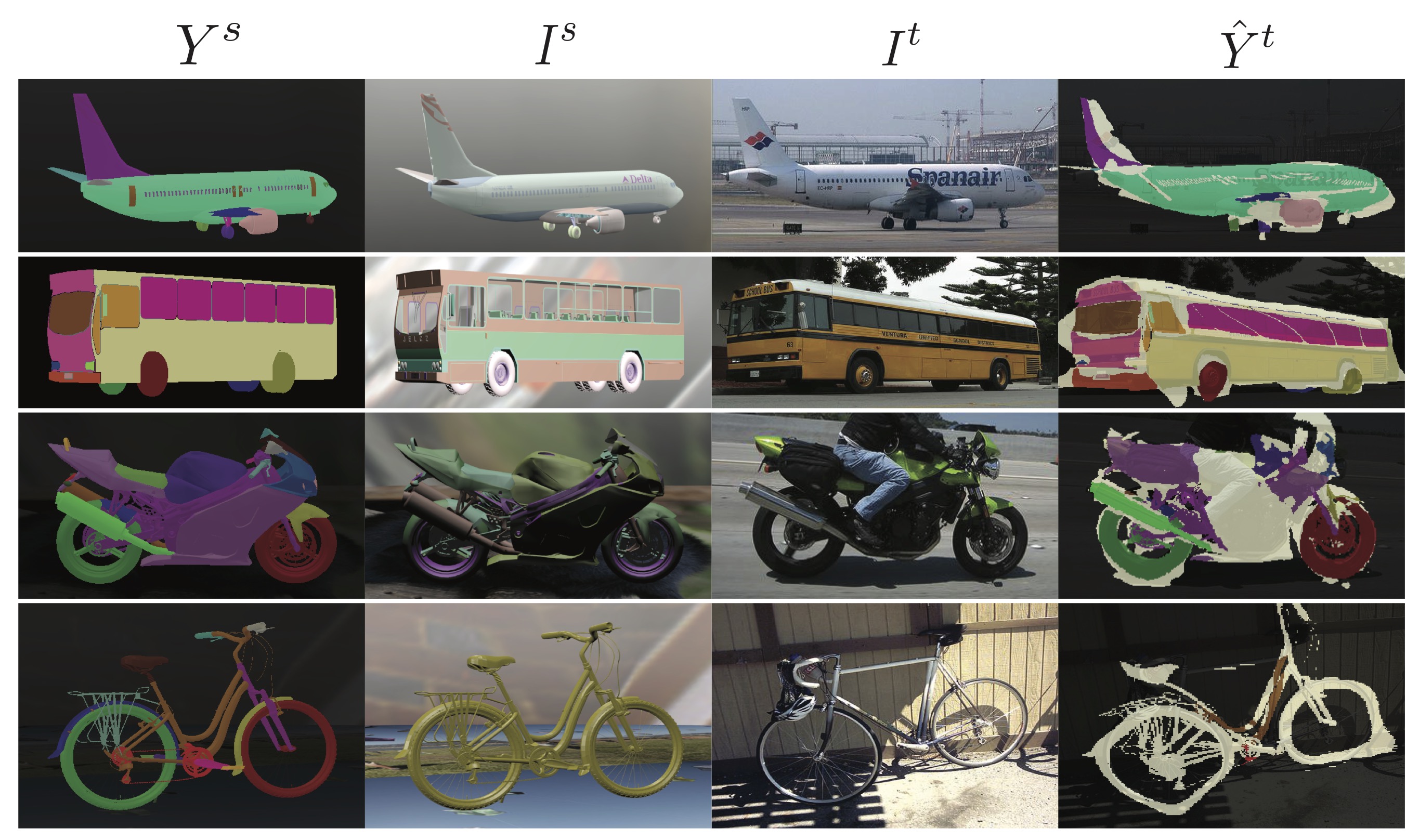}
\vspace{-0.3cm}
\caption{Quality of GMG pseudo-labels. Column 1-4 show the selected source label, the selected source image, the target image, and the final pseudo-label, respectively. Light yellow color indicates uncertain labels.}
\label{fig:GCG_rst}
\vspace{-0.5cm}
\end{figure}

\vspace{-0.1cm}

%% file: 05_conclusion.tex
\section{Conclusions}
\vspace{-0.2cm}
We introduce the idea of learning part segmentation for general objects from synthetic data through unsupervised domain adaptation (UDA). We first introduce UDA-Part, a comprehensive dataset for $5$ vehicle categories designed to benchmark part segmentation through UDA. Extending UDA-Part to more object categories is one of our future goal. Secondly, we propose a new UDA algorithm called Geometric-Matching Guided domain adaptation (GMG) which leverages the object's spatial structure to facilitate the adaptation process. In our experiments, GMG outperforms previous UDA methods on two real test image datasets, demonstrating the advantage of using structural information in UDA for part segmentation. On the other hand, GMG requires a grid-search process to find input pairs and is prone to misclassifying smaller parts, which could be improved in future works. In conclusion, our work provides a new solution for part segmentation on general objects with low cost and will motivate more research in this area.

\vspace{0.1cm}
\noindent
{\bf Acknowledgement}
This work was supported by NSF BCS-1827427, NIH R01 EY029700, and ONR N00014-21-1-2812. We thank Zihao Xiao for proofreading.

%% file: 06_supplementary.tex
\section{Supplementary Material}
\subsection{Part annotation tool for 3D CAD models}
\begin{figure}[h]
\centering
\includegraphics[width=0.8\linewidth]{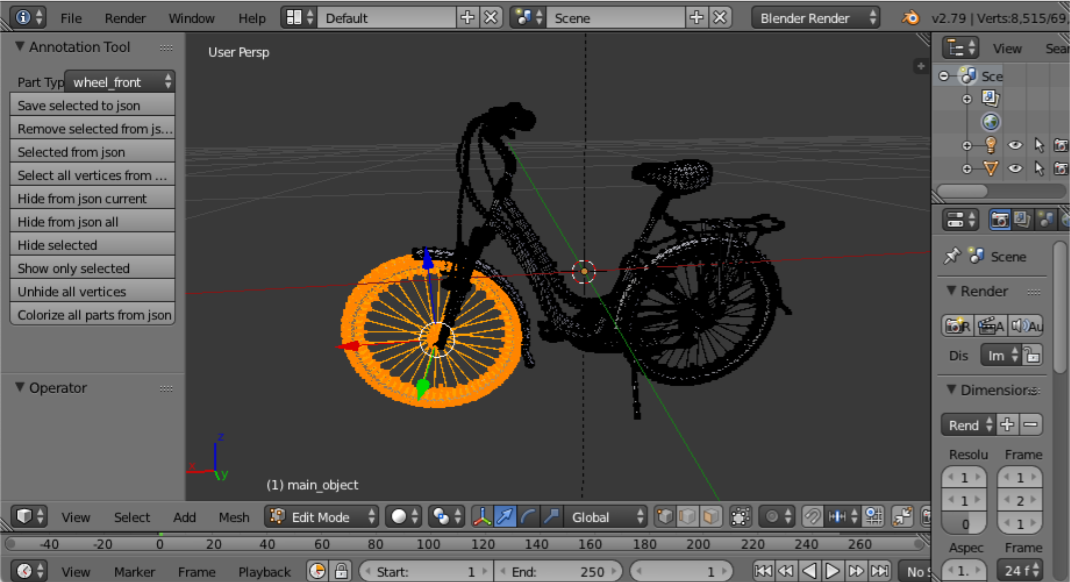}
\caption{A screenshot of the custom Blender plugin for per-mesh part labeling on 3D CAD models.}
\label{fig:blender}
\end{figure}

We use the Blender~\cite{blender} plugin built by Kim et al.~\cite{kim2022learn} to perform per-mesh part labeling on the 3D CAD models. A screen-shot of the software interface is shown in Figure~\ref{fig:blender}, where a group of meshes is selected and labeled as the part \emph{wheel\_front} for the \emph{bicycle} CAD model. 

\subsection{Part annotation tool for real test images}
\begin{figure}[h]
\centering
\includegraphics[width=0.8\linewidth]{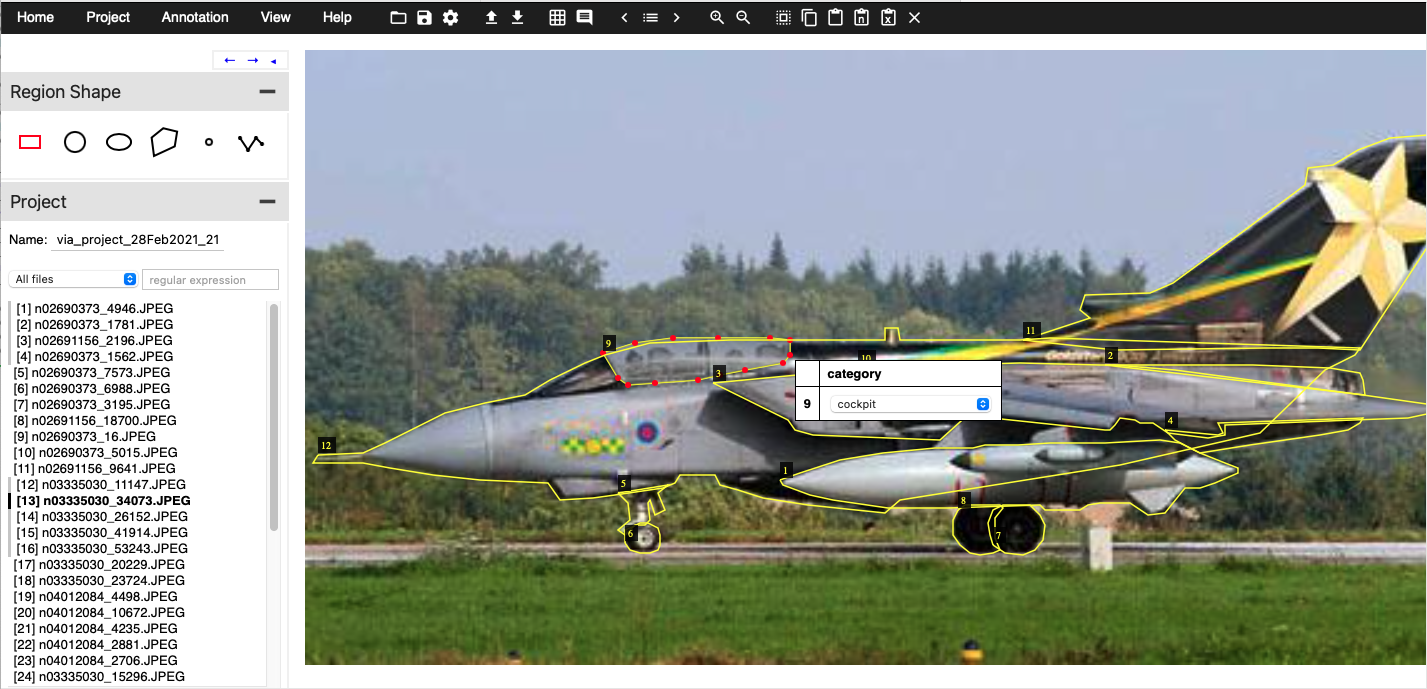}
\caption{A screenshot of VIA software for manual part annotation on real images.}
\label{fig:via}
\end{figure}

We use the VGG Image Annotator (VIA)~\cite{dutta2019vgg} to manually label the parts on the real images. A screen-shot of the annotator interface is shown in Figure~\ref{fig:via}, where vertices of a polygon are located to define the segmentation mask of the part \emph{cockpit} for the \emph{aeroplane} in this image.

\subsection{UDA-Part 3D CAD Models and Part Lists}
\begin{sidewaysfigure}[t]
\includegraphics[width=\textwidth]{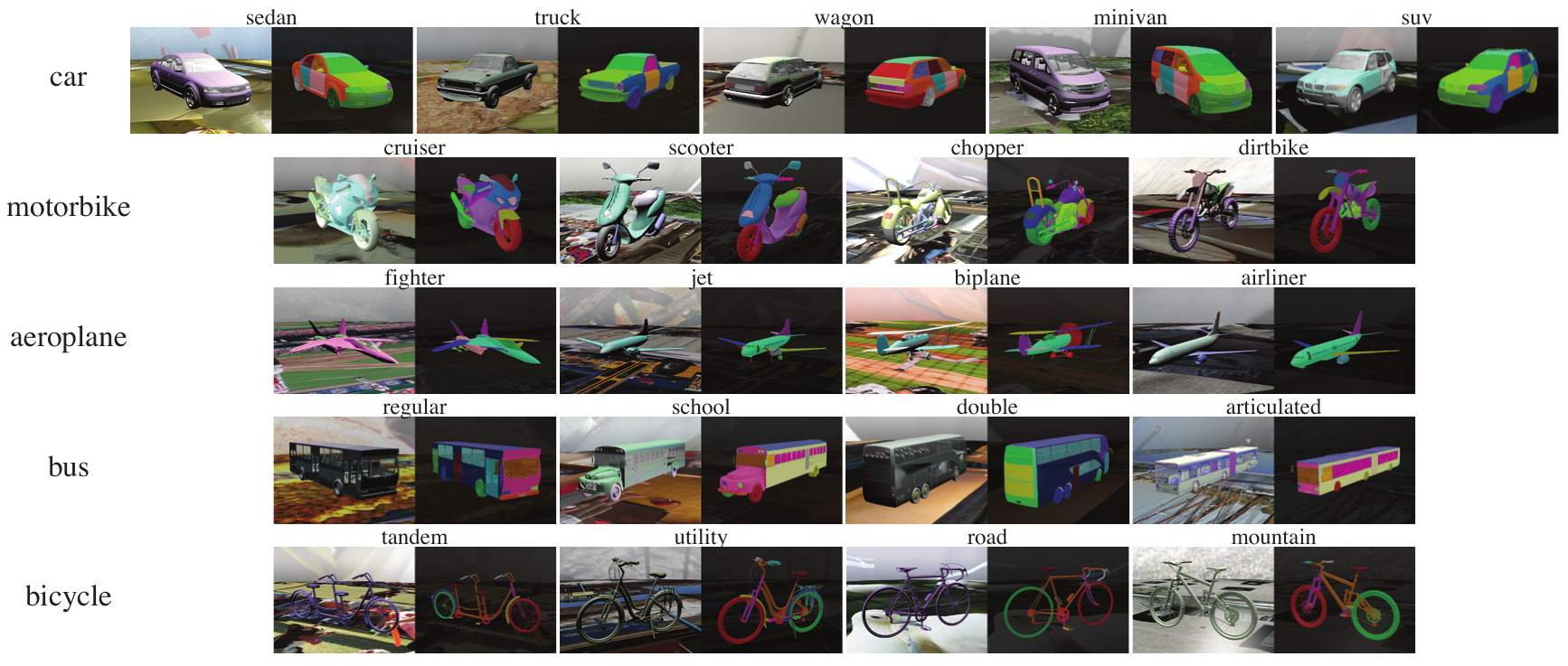}
\caption{3D CAD models/sub-categories included in UDA-Part.}
\label{fig:all_cad}
\end{sidewaysfigure}

UDA-Part is composed of $21$ 3D CAD models from $5$ vehicle categories, each with detailed per-mesh part labeling. Figure~\ref{fig:all_cad} shows the all $21$ CAD models and their corresponding part annotations. The full lists of parts annotated for each vehicle category can be found in Table~\ref{tab:list}.

\begin{table}[h]
\begin{tabular}{|l|l|}
\hline
category  & part list   \\ \hline
car       & \begin{tabular}[c]{@{}l@{}}bumper\_back, door\_back\_left, wheel\_back\_left, window\_back\_left, license\_plate\_back, door\_back\_right,\\ wheel\_back\_right, window\_back\_right, windshield\_back, bumper\_front, door\_front\_left, wheel\_front\_left,\\ window\_front\_left, license\_plate\_front, door\_front\_right, wheel\_front\_right, window\_front\_right, windshield\_front,\\ hood, frame\_left, head\_light\_left, mirror\_left, quarter\_window\_left, tail\_light\_left,\\ frame\_right, head\_light\_right, mirror\_right, quarter\_window\_right, tail\_light\_right, roof, trunk\end{tabular} \\ \hline
motorbike & \begin{tabular}[c]{@{}l@{}}wheel\_front, wheel\_back, fender\_front, fender\_back, frame, mirror\_left,\\ mirror\_right, windscreen, license\_plate, seat, seat\_back, gas\_tank,\\ handle\_left, handle\_right, headlight, taillight, exhaust\_left, exhaust\_right,\\ engine, cover\_front, cover\_body\end{tabular}  \\ \hline
aeroplane & \begin{tabular}[c]{@{}l@{}}propeller, cockpit, wing\_left, wing\_right, fin, tailplane\_left,\\ tailplane\_right, wheel\_front, landing\_gear\_front, wheel\_back\_left, landing\_gear\_back\_left, wheel\_back\_right,\\ landing\_gear\_back\_right, engine\_left, engine\_right, door\_left, door\_right, bomb\_left,\\ bomb\_right, window\_left, window\_right, body\end{tabular} \\ \hline
bus       & \begin{tabular}[c]{@{}l@{}}wheel\_front\_left, wheel\_front\_right, wheel\_back\_left, wheel\_back\_right, door\_front\_left, door\_front\_right,\\ door\_mid\_left, door\_mid\_right, door\_back\_left, door\_back\_right, window\_front\_left, window\_front\_right,\\ window\_back\_left, window\_back\_right, licplate\_front, licplate\_back, windshield\_front, windshield\_back,\\ head\_light\_left, head\_light\_right, tail\_light\_left, tail\_light\_right, mirror\_left, mirror\_right,\\ bumper\_front, bumper\_back, trunk, roof, frame\_front, frame\_back, frame\_left, frame\_right\end{tabular}    \\ \hline
bicycle   & \begin{tabular}[c]{@{}l@{}}wheel\_front, wheel\_back, fender\_front, fender\_back, fork, handle\_left,\\ handle\_right, saddle, drive\_chain, pedal\_left, pedal\_right, crank\_arm\_left,\\ crank\_arm\_right, carrier, rearlight, side\_stand, frame\end{tabular}  \\ \hline
\end{tabular}
\caption{Part list for each vehicle category in UDA-Part}
\label{tab:list}
\end{table}

\subsection{More detailed comparisons between UDA-Part and PascalPart~\cite{chen2014detect}}
\begin{figure}[t]
\centering
\includegraphics[width=0.95\linewidth]{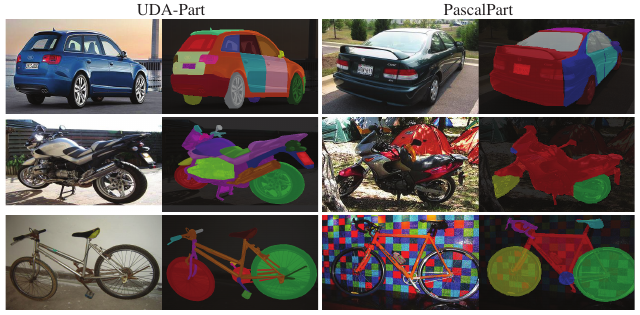}
\caption{Part annotation comparisons between UDA-Part and PascalPart~\cite{chen2014detect} dataset. Note UDA-Part provides more fine-grained part annotations.}
\label{fig:comp}
\end{figure}

In Table~\ref{tab:vspp}, we list out more details about UDA-Part and make per-category comparisons with PascalPart~\cite{chen2014detect}. The number of training samples in PascalPart is small, while UDA-Part provides large-scale synthetic images that is more adequate for deep neural network training. Also, the number of parts in UDA-Part is $2$ to $4$ times of the number in PascalPart, indicating UDA-Part provides more detailed part labeling for each category. In Figure~\ref{fig:comp}, we visualize several examples to compare the part annotations in UDA-Part and PascalPart. UDA-Part provides more fine-grained part annotations and these detailedly labeled parts will be more useful for various tasks.

\begin{figure}[t]
\centering
\includegraphics[width=0.8\linewidth]{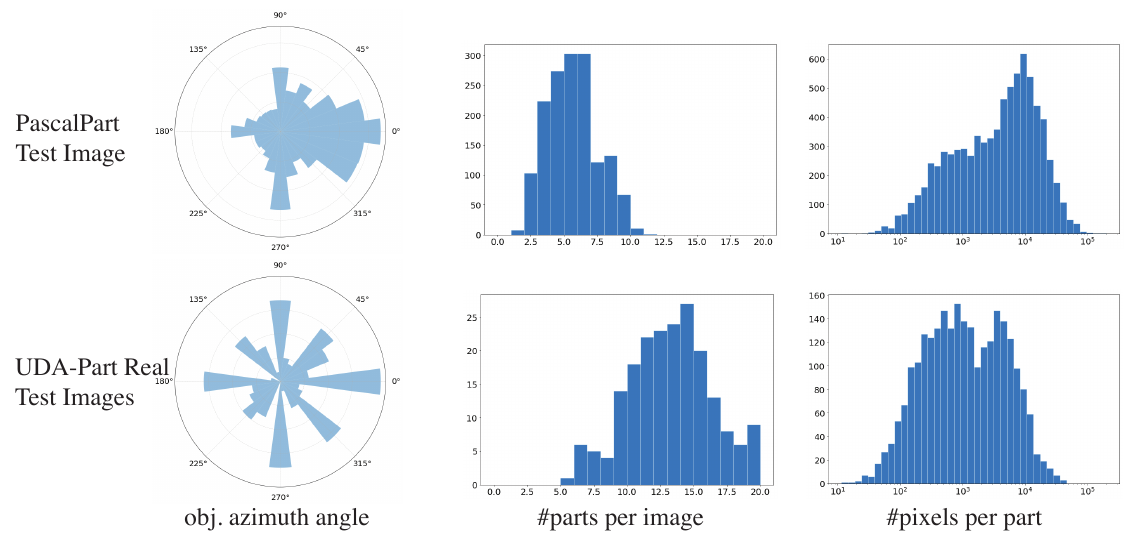}
\caption{Dataset statistics comparisons between UDA-Part and PascalPart~\cite{chen2014detect} dataset.}
\label{fig:stats}
\end{figure}

In Figure~\ref{fig:stats}, we further compare the distribution of object azimuth angles, number of parts per image, and number of pixels per part in PascalPart and UDA-Part real test images. PascalPart is biased to objects in the front-view, while UDA-Part contains more evenly distributed viewpoints, which results in more evenly distributed part instances. The number of parts per image in PascalPart is $2$ to $3$ times smaller than UDA-Part, which agrees with the fact that the number of parts labeled per category in PascalPart is also $2$ to $3$ times smaller. Moreover, the majority of parts in PascalPart have more than $5000$ pixels, while the parts in UDA-Part have evenly distributed sizes between $100$ to $10000$ pixels, indicating UDA-Part is a more challenging dataset for part segmentation.

\begin{table*}[h]
\centering
\begin{tabular}{|L{28mm}|L{18mm}|C{18mm}|C{18mm}|C{18mm}|C{18mm}|C{18mm}|}
\hline
\multicolumn{2}{|l|}{}                                             & car                                                                      & motorbike                                                                & aeroplane                                                                & bus                                                                      & bicycle                                                                  \\ \hline
\multirow{2}{*}{\# of training images}                       & UDA-Part    & (S) $30,000$ & (S) $24,000$ & (S) $24,000$ & (S) $24,000$ & (S) $24,000$ \\ \cline{2-7}
                                                    & PascalPart & (R) $538$                                                                 & (R) $261$                                                                  & (R) $266$                                                                  & (R) $221$                                                                  & (R) $252$                                                                  \\ \hline
\multirow{2}{*}{\# of test images}                       & UDA-Parts    & \begin{tabular}[c]{@{}c@{}}(S) $10,000$\\ (R) $40$\end{tabular} & \begin{tabular}[c]{@{}c@{}}(S) $8,000$\\ (R) $40$\end{tabular} & \begin{tabular}[c]{@{}c@{}}(S) $8,000$\\ (R) $40$\end{tabular} & \begin{tabular}[c]{@{}c@{}}(S) $8,000$\\ (R) $40$\end{tabular} & \begin{tabular}[c]{@{}c@{}}(S) $8,000$\\ (R) $40$\end{tabular} \\ \cline{2-7}
& PascalPart & (R) $520$ & (R) $255$ & (R) $280$ & (R) $229$ & (R) $263$  \\ \hline
\multirow{2}{*}{\# of parts}                        & UDA-Parts    & $31$          & $21$  & $22$   & $32$   & $17$    \\ \cline{2-7}
                                                    & PascalPart & $13$      & $6$  & $6$  & $13$   & $7$    \\ \hline
\end{tabular}
\caption{Per-category comparisons between UDA-Part and PascalPart~\cite{chen2014detect}. (S) indicates synthetic images, while (R) indicates real images. Note UDA-Part provides more samples which is more adequate for deep neural network training, and more number of parts per category which makes the segmentation task more challenging.}
\label{tab:vspp}
\end{table*}

\subsection{More training details for GMG.}
We implement GMG using Pytorch~\cite{NEURIPS2019_9015} on two TitanX GPUs. Synthetic training images are resized to have a long edge of $800$ pixels while real training images are resized to have a short edge of $224$. We apply random scaling between $0.5$ and $2$ and random crop of $513\times 513$ to all input samples. For evaluation, the real test images are resized to short edge $224$. We use batch size $12$ for training. The Source-Only model $\mathbf{M}^\mathcal{S}$ is trained for $50$ epochs using an SGD optimizer, with momentum set to $0.9$ and weight decay equals $1e-4$. The learning rate starts at $0.007$ and decreases every epoch using a polynomial scheduler with power $0.9$. For geometric matching, we use thin plate spline transformation with $25$ anchor points and take the first four convolutional blocks of an ImageNet~\cite{deng2009imagenet} pre-trained VGG16 network~\cite{simonyan2014very} as the feature extractor. The confidence threshold $\gamma$ is set to the $60$th percentile of the scores obtained from all samples in the corresponding category. We implement the pair selection using python with $8$ parallel CPU threads. For a pool of $24$ candidates, it takes $2.1$ seconds per image and roughly $54$ ($195$) minutes for the training set of PascalPart (PASCAL3D+). When the ground-truth viewpoint is given, the matching takes less than $0.4$ second per image, and thus roughly $10$ ($30$) minutes for PascalPart (PASCAL3D+). During joint training, we apply strong augmentations to synthetic images following~\cite{mu2020learning}. The joint training takes $10000$ iterations and the learning rate is fixed at $2.5e-4$. The real loss coefficient $\lambda$ is set to $1.0$ for all ``w/vp'' experiments and $0.1$ for all others.

\subsection{Potential of using UDA-Part for fully-supervised domain adaptation}
Besides unsupervised domain adaptation, UDA-Part can also be applied to fully-supervised domain adaptation by using PascalPart training labels. Here, we use na\"ive  fine-tuning approach to illustrate the potential. More specifically, we use PascalPart training samples with part labels to train two DeepLabv3+ models: one model is initialized from ImageNet pre-trained weights, and then trained on PascalPart; the other one is first trained for UDA-Part segmentation, then fine-tuned on PascalPart. The results are shown in Table~\ref{tab:fda}. Using UDA-Part pre-trained weights consistently improves the final performance on each category for more than $1$ point. UDA-Part pre-training helps the model to start with more meaningful deep features and benefits part segmentation in real image domain even when only na\"ive fine-tuning is used.

\begin{table}[]
\centering
\begin{tabular}{|L{20mm}|C{20mm}|C{20mm}|}
\hline
\multirow{2}{*}{} & \multicolumn{2}{c|}{Init. from} \\ \cline{2-3} 
                  & ImageNet        & UDA-Part       \\ \hline
 car               & $40.36$           & $41.59$         \\ \hline
motorbike         & $38.08$           & $39.42$         \\ \hline
aeroplane         & $42.47$           & $44.07$         \\ \hline
bus               & $34.42$           & $36.36$         \\ \hline
bicycle           & $40.57$           & $41.22$        \\ \hline
\end{tabular}
\caption{Comparison of weight initializations for fully supervised part segmentation in the real image domain. PascalPart~\cite{chen2014detect} training labels are used to train DeepLabv3+ models, and the performance (mIoU) on the test split is shown here. Note if the model is pre-trained on UDA-Part segmentation task, the final results could be consistently improved by more than $1$ point.}
\label{tab:fda}
\end{table}